\documentclass[twocolumn,10pt]{jmlr}

\usepackage{booktabs}
\usepackage{siunitx}
\usepackage[switch]{lineno}

\usepackage{float}
\restylefloat{figure}
\restylefloat{table}
\usepackage{enumitem}

\usepackage[T1]{fontenc}

\usepackage[most]{tcolorbox}
\tcbuselibrary{listings,skins,breakable}
\usepackage{xcolor}

\definecolor{PromptFrame}{RGB}{18,86,136}   
\definecolor{PromptBack}{RGB}{245,248,252}  
\definecolor{PromptTitle}{RGB}{15,23,42}    

\newtcblisting{promptbox}[2][]{%
  enhanced,
  breakable,
  colback=PromptBack,
  colframe=PromptFrame,
  coltitle=PromptTitle,
  fonttitle=\bfseries,
  boxrule=0.6pt,
  arc=1pt,
  outer arc=1pt,
  left=8pt,right=8pt,top=6pt,bottom=6pt,
  titlerule=0pt,
  title={#2},
  listing only,
  listing engine=listings,
  listing options={
    basicstyle=\ttfamily\footnotesize,
    breaklines=true,
    columns=fullflexible,
    keepspaces=true,
    showstringspaces=false,
    upquote=true
  },
  #1
}

\title[LLM for Evidence-Based Medical QA]{Evaluating Large Language Models for Evidence-Based Clinical Question Answering}

\author{
\Name{Can Wang} \Email{cwang271@jh.edu}\\
\addr Johns Hopkins University, USA
\AND
\Name{Yiqun Chen} \Email{yiqunc@jhu.edu}\\
\addr Johns Hopkins University, USA
}

\begin{document}

\maketitle

\begin{abstract}
Large Language Models (LLMs) have demonstrated substantial progress in biomedical and clinical applications, motivating rigorous evaluation of their ability to answer nuanced, evidence-based questions. We curate a multi-source benchmark drawing from Cochrane systematic reviews and clinical guidelines, including structured recommendations from the American Heart Association and narrative guidance used by insurers. Using GPT-4o-mini and GPT-5, we observe consistent performance patterns across sources and clinical domains: accuracy is highest on structured guideline recommendations (90\%) and lower on narrative guideline and systematic review questions (60--70\%). We also find a strong correlation between accuracy and the citation count of the underlying systematic reviews, where each doubling of citations is associated with roughly a 30\% increase in the odds of a correct answer. Models show moderate ability to reason about evidence quality when contextual information is supplied. When we incorporate retrieval-augmented prompting, providing the gold-source abstract raises accuracy on previously incorrect items to 0.79; providing top 3 PubMed abstracts (ranked by semantic relevance) improves accuracy to 0.23, while random abstracts reduce accuracy (0.10, within temperature variation). These effects are mirrored in GPT-4o-mini, underscoring that source clarity and targeted retrieval---not just model size---drive performance.   Overall, our results highlight both the promise and current limitations of LLMs for evidence-based clinical question answering. Retrieval-augmented prompting emerges as a useful strategy to improve factual accuracy and alignment with source evidence, while stratified evaluation by specialty and question type remains essential to understand current knowledge access and to contextualize model performance.
\end{abstract}


\paragraph*{Data and Code Availability}  
The raw materials are publicly available from their original repositories. We release our curated question–answer dataset, data processing scripts, and evaluation code at \url{https://github.com/yiqunchen/MEDAL}.


\section{Introduction}
Large Language Models (LLMs) have demonstrated strong capabilities in open-domain and medical question answering and reasoning~\citep{kamalloo_evaluating_2023,singhal_toward_2025}, but their performance in complex, evidence-based clinical domains remains an active area of exploration~\citep{singhal_toward_2025}. While prior benchmarks have evaluated biomedical QA performance across various formats~\citep{vladika_healthfc_2024,wan_what_2024,zheng_miriad_2025}, most existing datasets are derived from well-established medical practice and standardized questions (e.g., MedQA from medical licensing exams). The \emph{transportability} of these QA datasets to real-world clinical practice has recently been called into question~\citep{raji2025s,katz2024gpt}. This has spurred growing interest in whether LLMs can accurately address clinical questions grounded in diverse sources of evidence, particularly in settings that require reasoning about evidence quality. 

In particular, many clinically relevant questions are difficult to characterize because the underlying evidence may be missing or contradictory (e.g., differing results from clinical trials vs.\ observational studies). Moreover, such information is not readily available in standalone test formats (such as board exams), since the body of clinical evidence continuously evolves. A key avenue for capturing such evidence is through \emph{systematic reviews}, which are widely regarded as the gold standard for evidence-based medicine. Systematic reviews begin by surveying the full body of available research, then apply inclusion and exclusion criteria to filter eligible studies, extract relevant data (with graded evidence levels and risk-of-bias assessments), and finally synthesize findings—often using meta-analysis—to summarize the quantitative evidence. Increasingly, researchers have sought to leverage these rich textual and quantitative narratives to build more clinically realistic QA datasets~\citep{krithara_bioasq-qa_2023,vladika_medreqal_2024,polzak_can_2025}.

However, existing QA datasets are primarily designed to serve as benchmarks for LLMs and often fail to examine deeper characteristics of the underlying evidence (e.g., whether the cited studies are well-established or frequently cited, or the subject matter of the QA pair) and how these characteristics affect model accuracy. Moreover, most evaluations are structured as benchmarks of so-called “zero-shot” ability—where LLMs must answer without access to external tools for literature search or retrieval. 

To address these gaps, we construct a comprehensive, multi-source QA dataset to evaluate LLMs' ability to answer clinical questions and reason over supporting evidence. Our dataset includes questions derived from Cochrane systematic reviews,  and  structured recommendation guidelines from medical associations. With this diverse and carefully curated corpus, we aim to:  
\begin{enumerate}
    \item Assess the current performance of leading LLMs (as of August 2025, GPT-4o-mini and GPT-5 for small-scale and large-scale all-purpose language models, respectively).  
    \item Analyze trends such as LLMs tending to prefer answers supported by better-cited studies (controlling for model and publication year).  
    \item Evaluate LLMs in a retrieval-augmented generation (RAG) setting, where models can query PubMed as a proxy for web search, allowing us to measure how access to external evidence changes accuracy and reasoning.  
\end{enumerate}

\paragraph{Related Work.}
Biomedical and clinical QA research has been shaped by datasets such as PubMedQA~\citep{jin_pubmedqa_2019} and BioASQ-QA~\citep{krithara_bioasq-qa_2023}, which feature expert-annotated questions grounded in research abstracts. PubMedQA provides yes/no/maybe questions on research articles, while BioASQ-QA includes multi-format questions and summary answers, supporting both factual retrieval and summarization. More recent datasets such as MIRIAD~\citep{zheng_miriad_2025} scale to web collections, providing millions of QA pairs to enhance diversity and practical relevance.

HealthFC~\citep{vladika_healthfc_2024} examines the alignment between health claims and supporting/refuting evidence, annotated for veracity and strength. CONFLICTINGQA~\citep{wan_what_2024} collects controversial queries with conflicting evidence, showing that LLMs often prioritize surface relevance over deeper reasoning. MedREQAL~\citep{vladika_medreqal_2024} introduces QA pairs from Cochrane reviews, emphasizing recall and justification. MedEvidence~\citep{polzak_can_2025} directly compares LLM outputs against review conclusions, probing evidence synthesis. Beyond reviews, clinical guidelines provide structured recommendations based on evidence or consensus, and adherence is critical for decision-making~\citep{woolf_potential_1999}. Recent work has evaluated LLMs against these standards: MedGUIDE~\citep{li_medguide_2025} tests adherence to decision trees from guidelines, and AMEGA~\citep{fast_autonomous_2024} offers a broad benchmark spanning diagnosis, reasoning, and treatment planning.

Closest to our study are MedEvidence and MedREQAL, which both rely on systematic reviews. We extend them in two ways: (1) incorporating clinical guidelines, which more closely reflect real-world information channels and practice, and (2) providing a detailed performance breakdown to identify factors underlying stronger or weaker model performance.

\section{Methods}

\subsection{Dataset construction}
We constructed a multi-source clinical QA dataset from three evidence streams: Cochrane systematic reviews, American Heart Association (AHA) guideline recommendations, and narrative clinical guidelines. The Cochrane set included 8{,}533 abstracts of completed reviews (2010--2025, protocols excluded) with metadata (DOI, PubMed ID, title, abstract, authors, affiliations, year, citation counts)~\citep{CochraneLibrary}. AHA guidelines contributed 2{,}581 structured recommendations (2020--2025) with normalized Class of Recommendation (COR) and Level of Evidence (LOE)~\citep{AHAGuidelines}. Narrative guidelines consisted of 289 documents drawn from U.S. professional societies and major insurers.

Cochrane abstracts were enumerated via PubMed and Cochrane Library DOIs; metadata and structured abstracts were retrieved programmatically. For AHA guidelines, recommendations were extracted from machine-readable tables, appendices, and inline text. Narrative guidelines were collected from full-text or compiled sources, with duplicates and non-guideline content removed.

For dataset generation, GPT-4o produced structured question--answer (QA) labels across all sources. From {Cochrane abstracts}, we derived three types of outputs: (1) a clinically relevant question with one of (\{Yes, No, Not enough Evidence\}); (2) a question on whether the abstract reported discrepancies between findings from observational studies and randomized controlled trials (RCTs); and (3) a label for the overall quality of evidence, constrained to categorical values (five different levels).  

For {American Heart Association (AHA) recommendations}, we generated: (1) a judgment of whether the recommendation was supported by evidence (\{Yes, No, Unknown\}); (2) a rating of perceived recommendation strength; and (3) a rating of evidence quality. These labels were mapped directly to the AHA's guideline framework, which encodes both the strength and certainty of evidence. Specifically, the \emph{Class of Recommendation (COR)} indicates the strength of a clinical recommendation and has four categories: Class I (strong), Class IIa (moderate), Class IIb (weak), and Class III (no benefit or harm). The \emph{Level of Evidence (LOE)} indicates the type and quality of supporting evidence, with three main categories: Level A (high-quality evidence from multiple randomized trials or meta-analyses), Level B (moderate-quality evidence, including single randomized trials or nonrandomized studies), and Level C (expert opinion or limited data). 

Finally, for \textbf{narrative clinical guidelines}, each document was segmented into approximately 2{,}000-character chunks. From each chunk, GPT-4o generated a structured question specifying the Population, Intervention (or Exposure), Comparator, and Outcome. Answers were constrained to the categorical set \{Yes, No, No evidence\}.

To ensure quality and comparability, we harmonized COR/LOE categories across sources and required rationales to directly quote or closely paraphrase source text. Duplicate records were removed. For validation, we manually reviewed 100 Cochrane-derived QA pairs: all questions were relevant, 86\% were fully consistent, and 6\% partially correct due to ambiguous phrasing. Table~\ref{tab:dataset-examples} shows representative examples from each data source.

\begin{table}[h!] \floatconts {tab:dataset-examples} {\caption{Examples of generated questions and annotations across three data sources.}} { \scriptsize \setlength{\tabcolsep}{3.3pt} 
\begin{tabular}{p{2.4cm} p{4.9cm}} \toprule \multicolumn{2}{l}{\textbf{(a) Cochrane review abstract}} \\ \midrule 
\textbf{Title} & \textit{Chemoradiotherapy for cervical cancer: meta-analysis} \\ \textbf{Question} & Does chemoradiotherapy improve 5-year survival vs.\ radiotherapy alone in women with cervical cancer? \\ \textbf{Answer} & Yes \\ \textbf{Evidence} & High \\ \textbf{Discrepancy} & No \\ \textbf{Note} & 6\% improvement in 5-year survival (HR = 0.81, $p<0.001$). \\ \midrule \multicolumn{2}{l}{\textbf{(b) AHA guideline recommendation}} \\ \midrule \textbf{Recommendation} & For intermediate-risk patients with acute chest pain and no known coronary artery disease, a rest–stress myocardial perfusion imaging study is reasonable. \\ \textbf{Questions} & Q1: Supported by evidence? \newline Q2: Strength (1–5)? \newline Q3: Evidence quality (1–5)? \\ \textbf{Labels} & Class of Recommendation (COR): IIa; Level of Evidence (LOE): A \\ \textbf{Source} & Gulati et al., 2021 AHA/ACC Chest Pain Guideline \\ \midrule \multicolumn{2}{l}{\textbf{(c) Narrative clinical guideline}} \\ \midrule \textbf{Question} & In adults with chronic heart failure, does exercise therapy improve quality of life? \\ \textbf{Answer} & Yes \\ \textbf{Snippet} & Reports improved quality of life scores and physical capacity with structured exercise therapy. \\ \bottomrule \end{tabular} } \end{table}
\subsection{Evaluation}
We framed clinical QA as a three-way classification task with constrained label sets. To ensure consistency, models were prompted to produce structured outputs with a fixed set of fields (\texttt{question}, \texttt{answer}, \texttt{evidence-quality}, \texttt{discrepancy}, \texttt{notes}), each restricted to predefined values. The full templates used are provided in Appendix~\ref{appendix:prompt}.

We evaluated both \texttt{GPT-4o-mini} and \texttt{GPT-5} models to compare performance trends between smaller, fast-inference models and larger frontier reasoning models. Evaluation settings included: (1) a no-context baseline (question only), with additional analysis on a subsample of misclassified questions (performed for two of the three datasets); and (2) context ablations on a challenging subset where the baseline failed, tested with no context, the correct abstract, a random abstract, or PubMed top-3 (up to three retrieved abstracts concatenated with separators, \emph{excluding} the original article, mimicking a real-world application scenario).

\subsection{Statistical analysis and reporting}
Our primary metric was exact-match accuracy on the \texttt{answer} label:
\[
\mathrm{Accuracy} = \frac{\#\{\text{predicted answer} = \text{ground truth}\}}{\#\{\text{all evaluated items}\}}.
\]
Invalid or out-of-vocabulary outputs were included in the denominator and scored as incorrect. Additional metrics included response success rate (valid predictions), confusion matrices for \texttt{answer}, \texttt{evidence-quality}, and \texttt{discrepancy}, and 95\% confidence intervals via normal approximation. We further conducted stratified analyses by clinical field, publication year, and citation count. Correlations between citation count and accuracy were assessed using logistic regressions.

\section{Results}
We evaluated a baseline model (GPT-4o-mini) and an advanced model (GPT-5) on three distinct clinical question-answering tasks. Our findings show that while GPT-5 consistently outperforms the baseline, both models exhibit similar performance patterns. Performance is highest on structured data, varies significantly by evidence prominence and clinical domain, and improves substantially with contextual information.

\subsection{Performance on Systematic Reviews}
\label{result:sys_review}
\paragraph{Overall accuracy.}
On systematic review abstracts, GPT-5 outperformed GPT-4o-mini across all tasks by a small margin (2-6\%), with the largest gap in overall answer accuracy. Both models showed weaker performance on discrepancy detection and evidence-quality classification (Table~\ref{tab:sr_summary}).

\begin{table}[h!]
\centering
\caption{Performance on systematic review questions.}
\label{tab:sr_summary}
\begin{tabular}{l c c}
\hline
\textbf{Task} & \textbf{GPT-4o-mini} & \textbf{GPT-5} \\
\hline
Answer & 60.3\% & 67.8\% \\
Discrepancy & 57.0\% & 59.1\% \\
Evidence Quality & 32.1\% & 38.8\% \\
\hline
\end{tabular}
\end{table}

\begin{figure}[h!]
\centering
\includegraphics[width=0.95\linewidth]{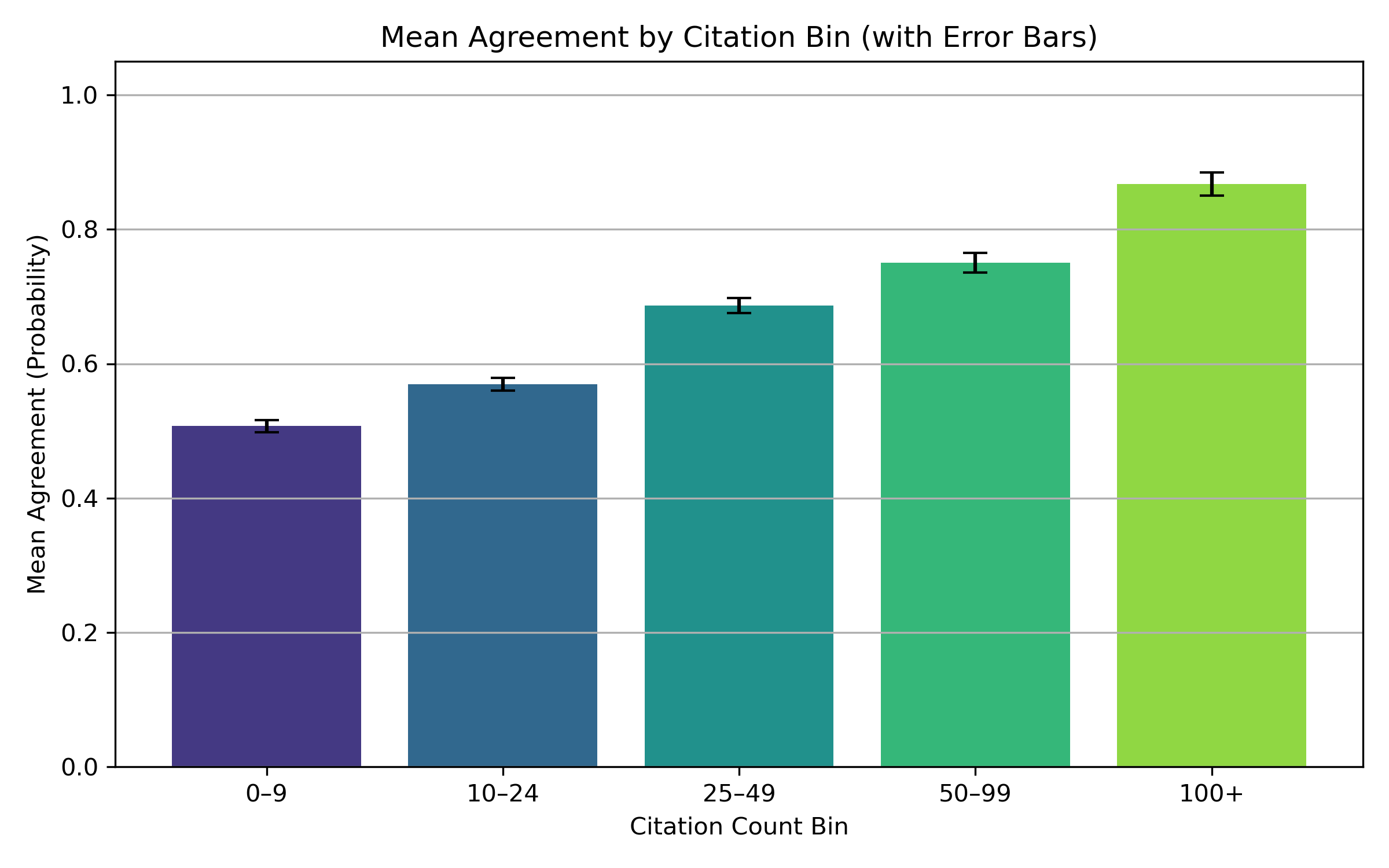}
\caption{Model answer accuracy by citation count bin with 95\% confidence intervals.}
\label{fig:citation-agreement}
\end{figure}

\begin{figure}[h!]
\centering
\includegraphics[width=0.95\linewidth]{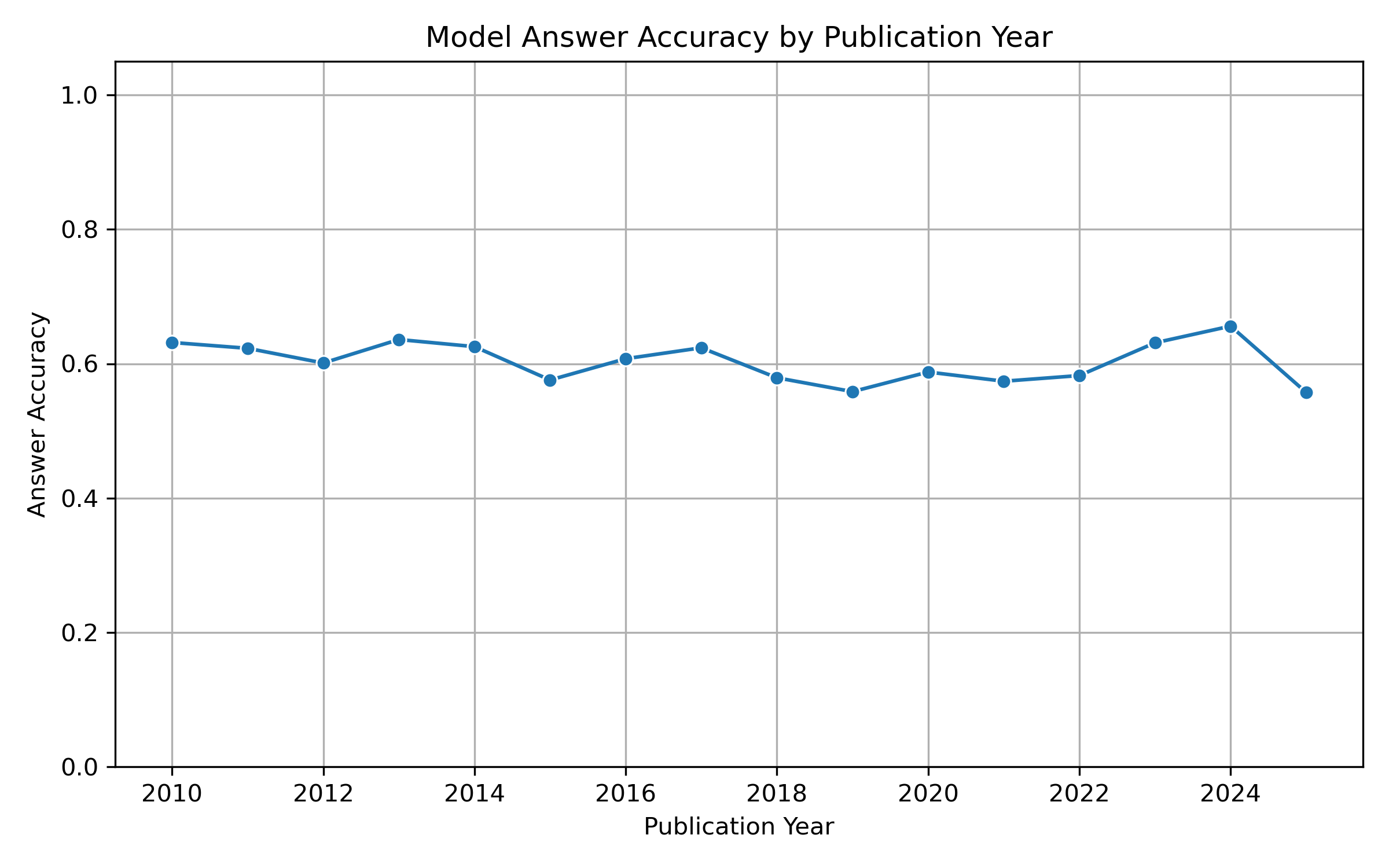}
\caption{Model answer accuracy by publication year with 95\% confidence intervals.}
\label{fig:accuracy-by-year}
\end{figure}

\paragraph{Influence of Evidence Prominence.}
Model accuracy was strongly associated with the citation prominence of the source review, indicating that 
findings from influential studies are more reliably encoded. For GPT-4o-mini, accuracy increased from about 
50\% for reviews with fewer than 10 citations to nearly 80\% for those with more than 100 (p$<$0.001). GPT-5 
showed the same trend, rising from 59.1\% in the lowest citation bracket to 79.1\% in the highest (odds ratio 
for $\log(1+\text{citations}) = 1.34$, 95\% CI: 1.29–1.40; Figure~\ref{fig:citation-agreement}). To test whether 
this effect was simply explained by older papers having more time to accumulate citations, we examined accuracy 
by publication year. Performance remained relatively stable between 2010 and 2015 (55–65\%) and did not increase 
monotonically with paper age; in fact, there was a slight decline for reviews published in 2025, likely reflecting 
the temporal cut-off of model training data (Figure~\ref{fig:accuracy-by-year}). Together, these results suggest 
that the observed citation effect reflects the prominence and impact of the underlying research rather than the 
age of the publication.

\paragraph{Variability Across Research Domains.}
We classified each systematic review into one of 37 primary research areas defined by the Cochrane classification. Performance varied considerably across domains for both models, with no strong correlation between the number of articles published in the primary research area and accuracy ($r = -0.14$). GPT-4o-mini achieved its highest accuracy in Rheumatology (72.7\%) and lowest in Wounds (43.4\%). GPT-5 showed a similar pattern, performing best in Tobacco, Drugs and Alcohol (74.6\%) and worst in Health and Safety at Work (52.1\%). Table~\ref{tab:topic-match-extremes} summarizes the top and bottom five domains for GPT-4o-mini (and corresponding GPT-5 model accuracy).
\begin{table}[h!]
\centering
\caption{Top and bottom 5 topics by answer accuracy (\%): GPT-4o-mini vs.\ GPT-5.}
\label{tab:topic-match-extremes}
\renewcommand{\arraystretch}{1.1}
\resizebox{\columnwidth}{!}{%
\begin{tabular}{l c c c}
\toprule
\textbf{Topic} & \textbf{Count} & \textbf{GPT-4o-mini} & \textbf{GPT-5} \\
\midrule
\multicolumn{4}{c}{\textit{Top 5 Topics}} \\
\midrule
Rheumatology & 66 & 72.7\% & 70.8\% \\
Pain and Anaesthesia & 337 & 68.8\% & 72.1\% \\
Tobacco, Drugs and Alcohol & 169 & 68.6\% & 74.6\% \\
Public Health & 108 & 68.5\% & 65.7\% \\
Urology & 108 & 67.6\% & 66.7\% \\
\midrule
\multicolumn{4}{c}{\textit{Bottom 5 Topics}} \\
\midrule
Health and Safety at Work & 48 & 54.2\% & 52.1\% \\
Complementary \& Alt. Med. & 92 & 53.3\% & 57.6\% \\
Dentistry and Oral Health & 216 & 53.2\% & 61.1\% \\
Neonatal Care & 400 & 50.5\% & 68.8\% \\
Wounds & 173 & 43.4\% & 61.8\% \\
\bottomrule
\end{tabular}
}
\end{table}
\paragraph{Error Analysis.}
A detailed error analysis of GPT-5 indicates strong performance on affirmative judgments but persistent
weakness on “No evidence” cases. As summarized in Table~\ref{tab:classification-report-gpt5}, GPT-5 achieves
its highest F1 on “Yes” answers (F1=0.80; precision=0.84, recall=0.76; support=5167), with notably lower F1
on “No” (0.61; support=2309) and especially “No evidence” (0.33; support=1054). For discrepancy detection,
GPT-5 shows good specificity for “No” (F1=0.70) but low sensitivity for “Yes” discrepancy (F1=0.13;
Table~\ref{tab:classification-report-gpt5}). This pattern suggests the model tends to produce confident,
plausible responses and under-calls uncertainty or conflicting evidence. A representative error case is shown
in Table~\ref{tab:error-antiplatelet}, where GPT-4o-mini generated a fluent but factually incorrect rationale
that contradicted high-quality trial evidence. For comparison, the GPT-4o-mini classification report—showing
even more pronounced difficulty on “No evidence” and discrepancy “Yes”—is provided in Appendix
Table~\ref{tab:classification-report-mini}.
\begin{table}[h!]
\centering
\caption{Classification report for GPT-5. Metrics derived from merged predictions and ground truth.}
\label{tab:classification-report-gpt5}
\resizebox{\columnwidth}{!}{%
\begin{tabular}{lcccc}
\toprule
\textbf{Answer} & \textbf{Precision} & \textbf{Recall} & \textbf{F1-score} & \textbf{Support} \\
\hline
No           & 0.58 & 0.65 & 0.61 & 2309 \\
No Evidence  & 0.30 & 0.37 & 0.33 & 1054 \\
Yes          & 0.84 & 0.76 & 0.80 & 5167 \\
\bottomrule
\end{tabular}
}
\vspace{0.5em}
\resizebox{\columnwidth}{!}{%
\begin{tabular}{lcccc}
\toprule
\textbf{Discrepancy} & \textbf{Precision} & \textbf{Recall} & \textbf{F1-score} & \textbf{Support} \\
\hline
Missing & 0.48 & 0.51 & 0.49 & 2756 \\
No      & 0.76 & 0.64 & 0.70 & 5544 \\
Yes     & 0.08 & 0.33 & 0.13 & 230 \\
\bottomrule
\end{tabular}
}
\end{table}

\begin{table}[h!]
\caption{Error example: GPT-4o-mini on Antiplatelet Agents.}
\label{tab:error-antiplatelet}
\small
\begin{tabular}{p{0.35\columnwidth} p{0.6\columnwidth}}
\toprule
\textbf{Question} & Do antiplatelet agents reduce all-cause mortality in patients with intermittent claudication compared to placebo? \\
\textbf{Model Answer} & No \\
\textbf{Model Notes} & Current evidence...suggests that they do not significantly reduce all-cause mortality... \\
\textbf{Ground Truth} & Yes \\
\textbf{GT Notes} & Antiplatelet agents reduced all-cause mortality with a risk ratio of 0.76 (95\% CI 0.60 to 0.98). \\
\bottomrule
\end{tabular}
\end{table}
\subsection{Performance on Structured Clinical Guidelines (AHA)}

When evaluated on highly structured recommendations from AHA guidelines, both models performed exceptionally well. GPT-4o-mini achieved 94.0\% accuracy. Importantly, its errors were not randomly distributed but concentrated in cases where the guidelines themselves provided weaker evidentiary support. Specifically, incorrect predictions clustered in recommendations with Level of Evidence (LOE) C-LD (Limited Data) or C-EO (Expert Opinion), and in Class of Recommendation (COR) 2B, which represents weak positive recommendations. In contrast, errors were rare for recommendations backed by strong trial evidence (LOE A) or those classified as unequivocal benefit or harm (Class 1 and Class 3). This indicates that the model’s uncertainty mirrors the ambiguity present in clinical evidence.

\begin{figure*}[htbp]
\centering
\includegraphics[width=0.9\textwidth]{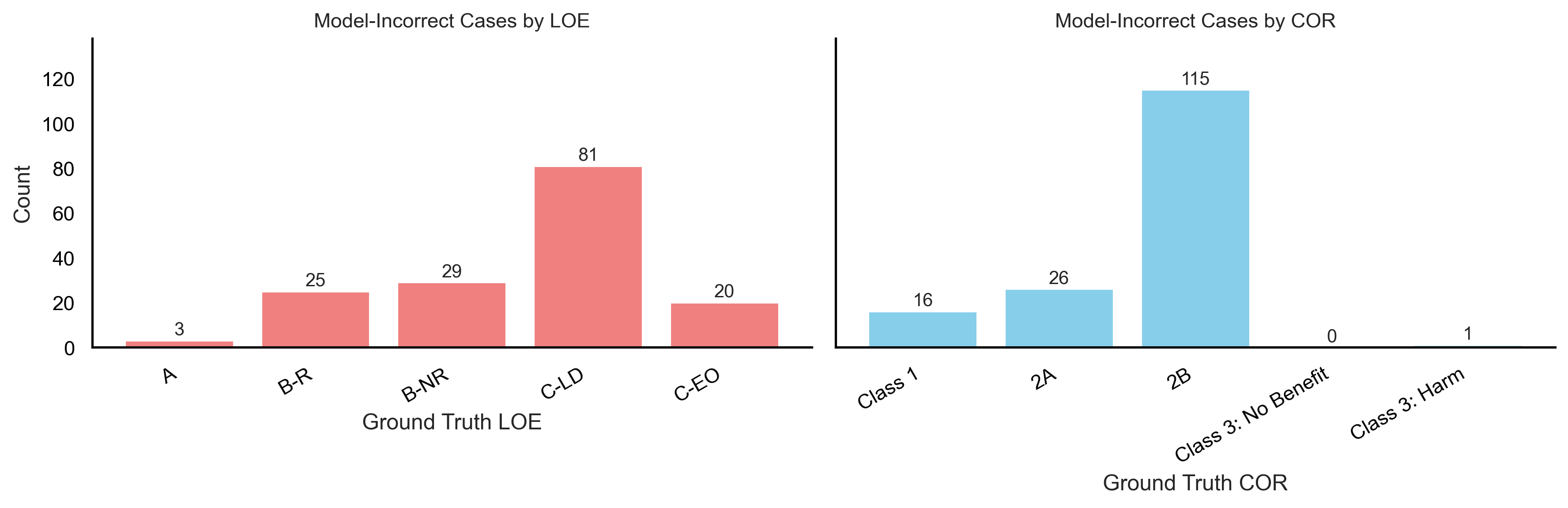}
\caption{Distribution of model-incorrect cases across guideline-defined evidence levels (LOE, left) and recommendation strengths (COR, right). Errors cluster in weak or low-confidence categories (C-LD, C-EO, Class 2B).}
\label{fig:aha-errors}
\end{figure*}

Figure~\ref{fig:aha-errors} illustrates this pattern. On the left, the majority of misclassifications occur in LOE C-LD (81 cases) and C-EO (20 cases), compared to very few in LOE A (3 cases). On the right, nearly all COR-related errors fall in Class 2B (115 cases), while Classes 1 and 3 (harm/no benefit) show almost no errors. These results highlight that the model is more likely to struggle when guideline recommendations themselves are based on limited or conflicting evidence.

\begin{figure*}[h]
\centering
\includegraphics[width=0.92\textwidth]{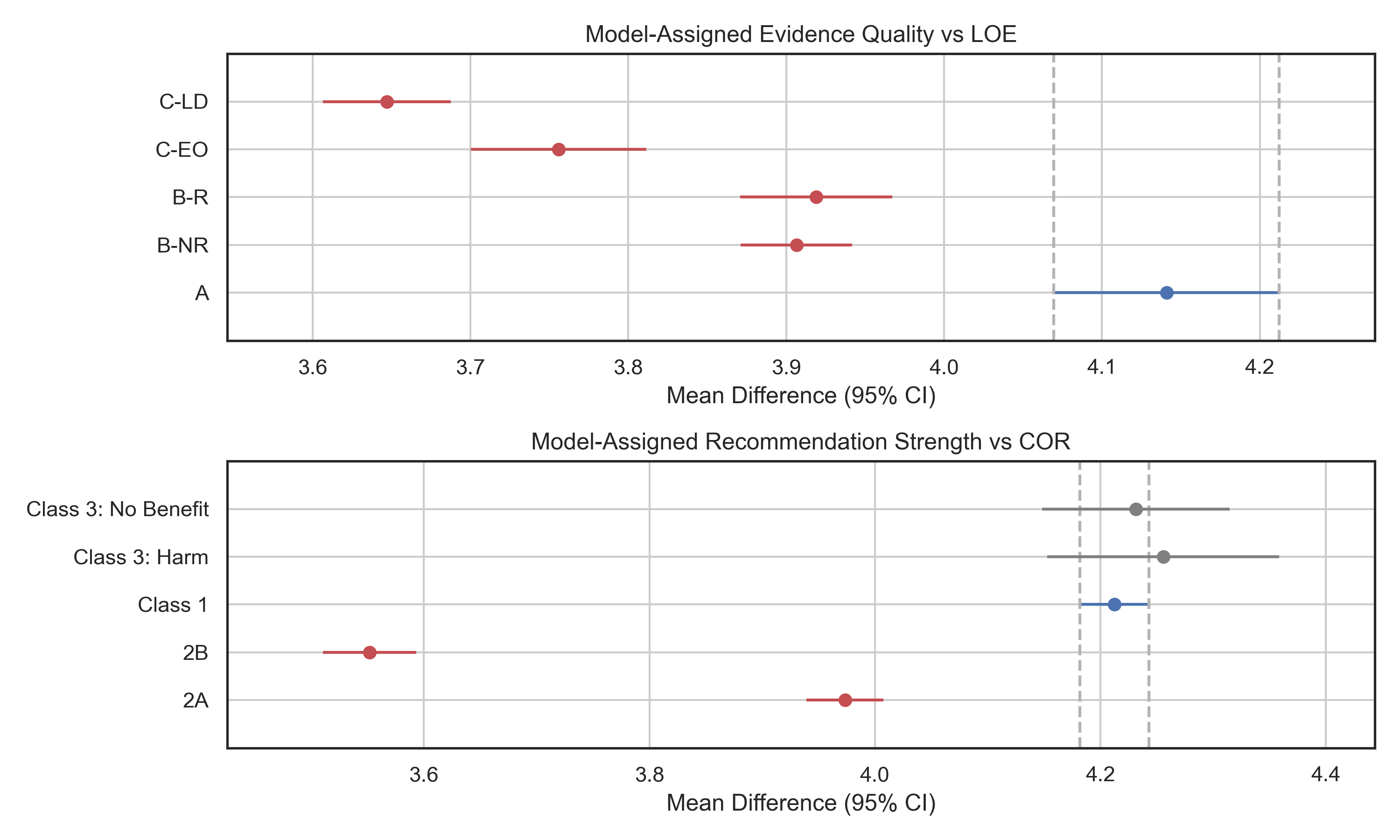}
\caption{Tukey HSD analysis of model-assigned scores versus ground-truth evidence levels (LOE, top) and recommendation strengths (COR, bottom). Higher ratings correspond to stronger evidence and recommendations, while weaker categories (C-LD, C-EO, Class 2B) received the lowest scores.}
\label{fig:aha-tukey}
\end{figure*}

Beyond raw accuracy, we examined whether the model’s assigned evidence-quality and recommendation-strength scores aligned with the clinical hierarchy. Figure~\ref{fig:aha-tukey} presents a Tukey HSD analysis comparing model-assigned scores against guideline-defined categories. For evidence quality (top panel), the model assigned the highest scores to LOE A (around 4.1–4.2 on a 1–5 scale), intermediate scores to B-R and B-NR (around 3.9), and the lowest scores to C-LD and C-EO (3.6–3.8). Statistical testing confirmed significant differences between most LOE categories, except between B-R and B-NR, which the model did not clearly separate. 

For recommendation strength (bottom panel), the model produced the highest ratings for strong recommendations: Class 1 (strong positive) and Class 3 (harm or no benefit) both received mean scores above 4.0. Class 2A (moderate positive) was rated slightly lower (just under 4.0), while Class 2B (weak positive) received the lowest ratings (around 3.6). This monotonic trend indicates that the model not only achieves high classification accuracy but also assigns scores that reflect the underlying gradient of confidence in the guidelines. In other words, the model’s internal scoring mirrors the clinical semantics of “strong” versus “weak” recommendations, with uncertainty concentrated in the same categories where clinicians themselves recognize ambiguity.

\subsection{Performance on Narrative Clinical Guidelines}
\begin{figure}[htbp!]
\centering
\includegraphics[width=0.45\textwidth]{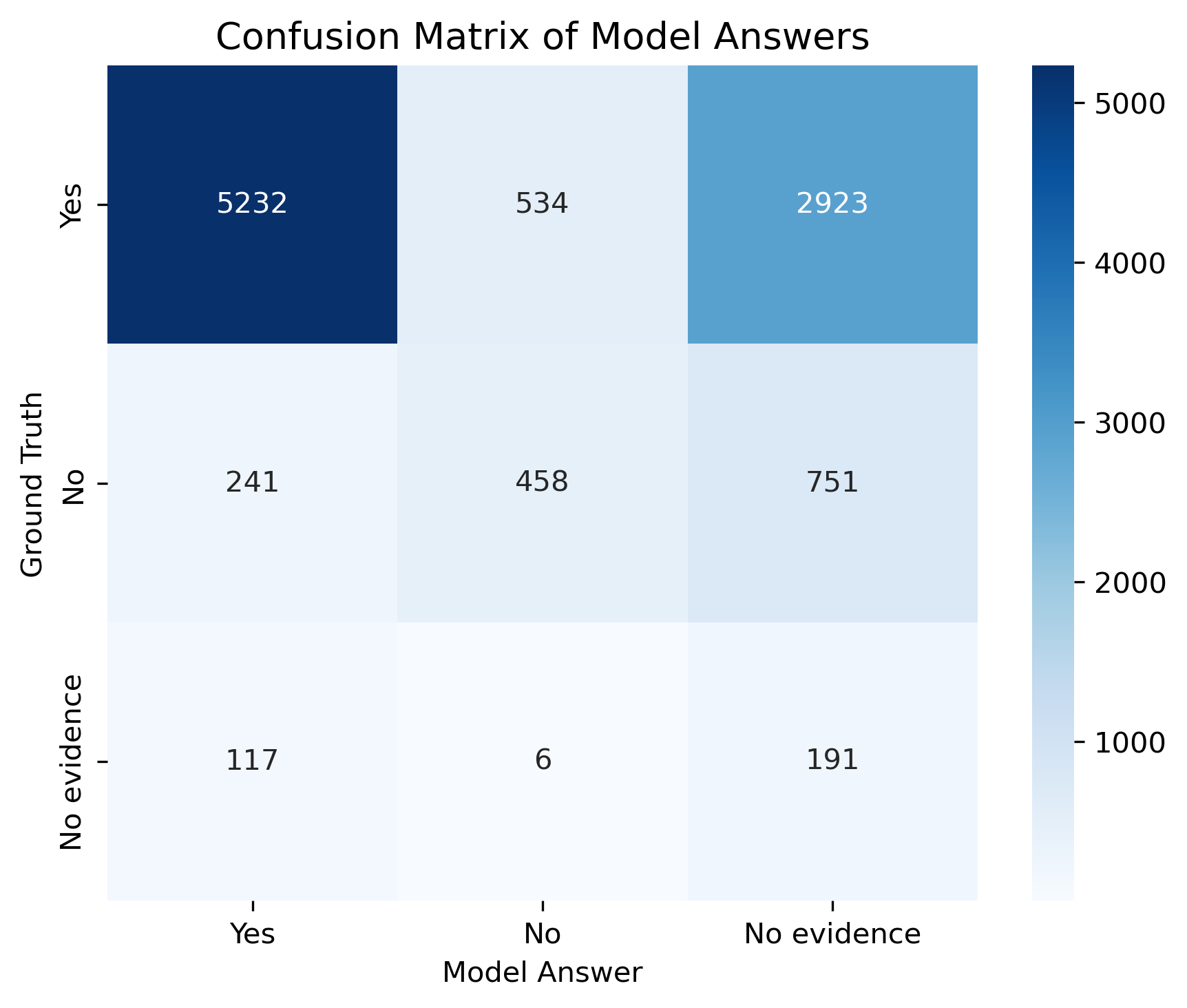}
\caption{Confusion matrix of GPT-4o-mini predictions on narrative-guideline-based questions.}
\label{fig:confusion-matrix-narrative}
\end{figure}

In  contrast to the structured AHA task, model accuracy dropped significantly on questions from unstructured, narrative text. GPT-4o-mini's accuracy fell to 56.3\%, where it particularly failed to correctly interpret statements of negative findings (31.6\% accuracy on "No" answers), as shown in Figure~\ref{fig:confusion-matrix-narrative}. This issue was especially pronounced with sentences containing double negations or complex phrases indicating a lack of efficacy. For instance, as shown in Table~\ref{tab:narrative-error}, the model incorrectly answered "Yes" to a question where the source text explicitly stated the intervention "provides no added benefit," demonstrating a bias towards affirmative responses when faced with nuanced negative language.

\begin{table}[htbp!]
\centering
\caption{Example of a model error on a narrative guideline QA.}
\label{tab:narrative-error}
\small
\begin{tabular}{p{0.22\columnwidth} p{0.68\columnwidth}}
\toprule
\textbf{Question} & In patients with chronic neck pain, does dry needling combined with guideline-based physical therapy provide additional benefit? \\
\textbf{Ground Truth} & No \\
\textbf{Model Answer} & Yes \\
\textbf{Supporting Snippet} & Dry needling combined with guideline-based physical therapy provides no added benefit... \\
\bottomrule
\end{tabular}
\end{table}

\subsection{Impact of Retrieval-Augmented Context}
To evaluate whether external context improves model performance, we re-ran the systematic review task 
(Section~\ref{result:sys_review}) under different contextual conditions. As summarized in 
Table~\ref{tab:context_summary}, providing the correct source abstract increased accuracy 
for both models, surpassing 90\%. A more realistic setup using PubMed-retrieved abstracts also yielded 
substantial gains, boosting GPT-4o-mini from a baseline of 60.3\% to 79.9\% and GPT-5 from 67.8\% to 
75.2\% on the tested subset. In contrast, irrelevant context (random abstracts) produced only a slight 
degradation in accuracy, suggesting both models are relatively robust to noisy inputs. These results 
highlight retrieval-augmented prompting as a practical and effective strategy for improving performance 
on systematic review–based clinical QA.

\begin{table}[htbp!]
\centering
\caption{Model accuracy on abstract-based QA with different contexts; *: GPT-5 contextual results are extrapolated from a 500-case subset of previously incorrect answers.}
\label{tab:context_summary}
\resizebox{\columnwidth}{!}{%
\begin{tabular}{l c c}
\toprule
\textbf{Context Condition} & \textbf{GPT-4o-mini} & \textbf{GPT-5*} \\
\midrule
No Context (Baseline)   & 60.3\% & 67.8\% \\
Correct Abstract (Gold) & 91.6\% & 93.2\% \\
PubMed Retrieval        & 79.9\% & 75.2\% \\
Random Abstract (Noise) & 58.1\% & 65.1\% \\
\bottomrule
\end{tabular}
}
\vspace{1mm}
\end{table}

\section{Discussion and Conclusion}
We present a new dataset and accompanying in-depth analysis for evaluating LLMs in evidence-based clinical question answering, grounded in high-quality sources including systematic reviews and clinical guidelines. Compared with existing biomedical QA datasets such as PubMedQA~\citep{jin_pubmedqa_2019}, BioASQ-QA~\citep{krithara_bioasq-qa_2023}, and more recent large-scale collections like MIRIAD~\citep{zheng_miriad_2025}, our dataset uniquely emphasizes reasoning about evidence quality, recommendation strength, and discrepancies between study designs. This focus aligns with recent calls for benchmark tasks that move beyond surface-level fact retrieval toward assessing how models handle conflicting or uncertain evidence~\citep{wan_what_2024, vladika_healthfc_2024}.

Our evaluation reveals several key findings. First, base models achieve only moderate accuracy on systematic review–derived questions, reflecting common error modes such as gaps in domain-specific knowledge and failures when evidence is absent or ambiguous. Second, performance is markedly higher on structured recommendations from clinical guidelines, consistent with prior observations that LLMs excel on templated biomedical tasks~\citep{sun2025generating}. Third, even without explicit context, models show some ability to differentiate the underlying evidence quality and recommendation strength of unstructured guideline recommendations. By contrast, evidence quality is much harder to recover from free-text systematic reviews or narrative guideline statements, where double negation and hedging language likely introduce substantial error. Fourth, performance varies systematically by domain and citation impact of the source literature, suggesting that model predictions are influenced not only by input complexity but also by the external visibility of the underlying evidence. Finally, in-context learning and retrieval-augmented prompting substantially improve performance, consistent with broader findings in biomedical retrieval-augmented generation~\citep{xiong2024benchmarking,matsumoto2024kragen}.

Our findings should be interpreted with the following limitations. First, a large proportion of our QA pairs were automatically generated via LLMs. Although manual spot-checking confirmed high consistency with source documents, errors and ambiguities may still be present. Second, our data sources, while high quality, are not exhaustive. Since we focused on Cochrane reviews and major guidelines, we excluded other specialties and grey literature, which may contain emerging evidence but have not yet risen to the level of systematic review. Given our emphasis on established high-quality evidence, this reflects a deliberate trade-off between timeliness and methodological rigor. Third, answer formats were mostly restricted to categorical labels to improve the efficiency of evaluation, which may oversimplify complex reasoning processes and preclude nuanced justifications.

Future work should extend this dataset to cover additional evidence sources, incorporate richer answer structures, and pursue more rigorous human validation. In parallel, integrating retrieval pipelines with high-quality biomedical databases and guideline repositories may offer a practical path toward safer, more robust clinical decision support systems powered by LLMs.

\bibliography{references}
\appendix
\onecolumn
\section{Prompt Templates}
\label{appendix:prompt}
This appendix lists the exact prompts used in the evaluation. Placeholders such as \texttt{\{question\}} are programmatically substituted. For context runs, retrieved abstracts are inserted verbatim under the “Background context” header. When multiple PubMed abstracts are present, they are concatenated with a separator line \texttt{---}.

\subsection{No-Context Prompt}
\begin{promptbox}{No-Context Prompt}
You are a clinical research expert with knowledge of systematic reviews, RCTs, and observational studies.
Task: Given a clinical question, return a JSON with keys \texttt{question}, \texttt{answer}, \texttt{evidence-quality}, \texttt{discrepancy}, \texttt{notes}.

Allowed values:
- answer: Yes | No | No Evidence
- evidence-quality: High | Moderate | Low | Very Low | Missing
- discrepancy: Yes | No | Missing

Question:
"""{question}"""
\end{promptbox}

\subsection{Context Prompt (scripts/evaluate\_with\_context.py)}
\begin{promptbox}{Context Prompt (with optional abstracts)}
You are a clinical research expert with knowledge of systematic reviews, RCTs, and observational studies.
Task: Given a clinical question, optionally with background abstracts, return a JSON with keys \texttt{question},
\texttt{answer}, \texttt{evidence-quality}, \texttt{discrepancy}, \texttt{notes}.

Allowed values:
- answer: Yes | No | No Evidence
- evidence-quality: High | Moderate | Low | Very Low | Missing
- discrepancy: Yes | No | Missing

Background context (may be empty; if multiple abstracts, separate with a line containing only \texttt{---}):
"""{context}"""

Question:
"""{question}"""
\end{promptbox}

\subsection{Notes on Inference Configuration}
\begin{itemize}
  \item Temperature set to \texttt{0.2} for non-reasoning models; for \texttt{GPT-5} we use \texttt{temperature=None} and, when supported, \texttt{reasoning\_effort=medium}.
  \item Responses requested as strict JSON via \texttt{response\_format=\{type=json\_object\}} when available; outputs are parsed with JSON deserialization. Invalid outputs are recorded as errors and counted as incorrect.
\end{itemize}

\section{Additional Results}

\begin{figure*}[h!]
\centering
\includegraphics[width=0.95\textwidth]{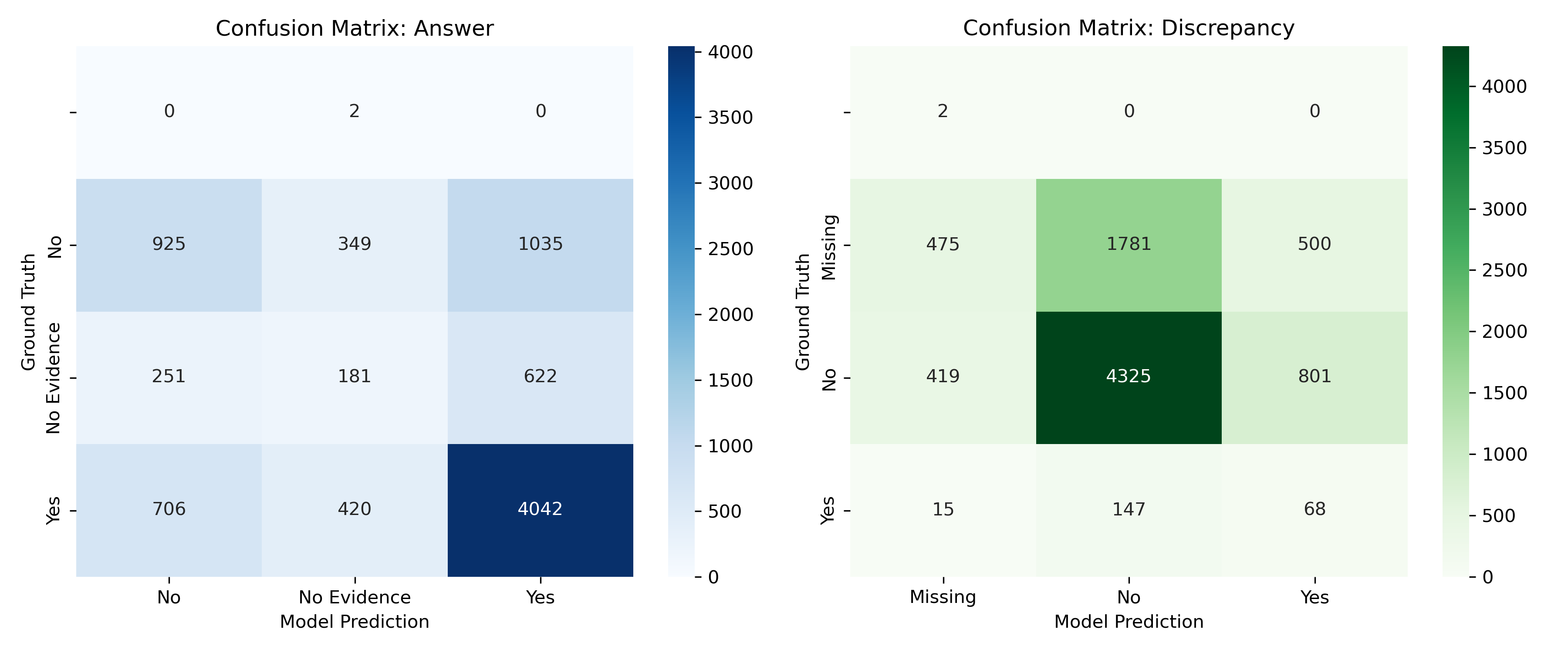}
\caption{Confusion matrices for answer prediction and discrepancy detection (GPT-4o-mini).}
\label{fig:confusion-matrix}
\end{figure*}

\begin{figure}
    \centering
    \includegraphics[width=0.5\linewidth]{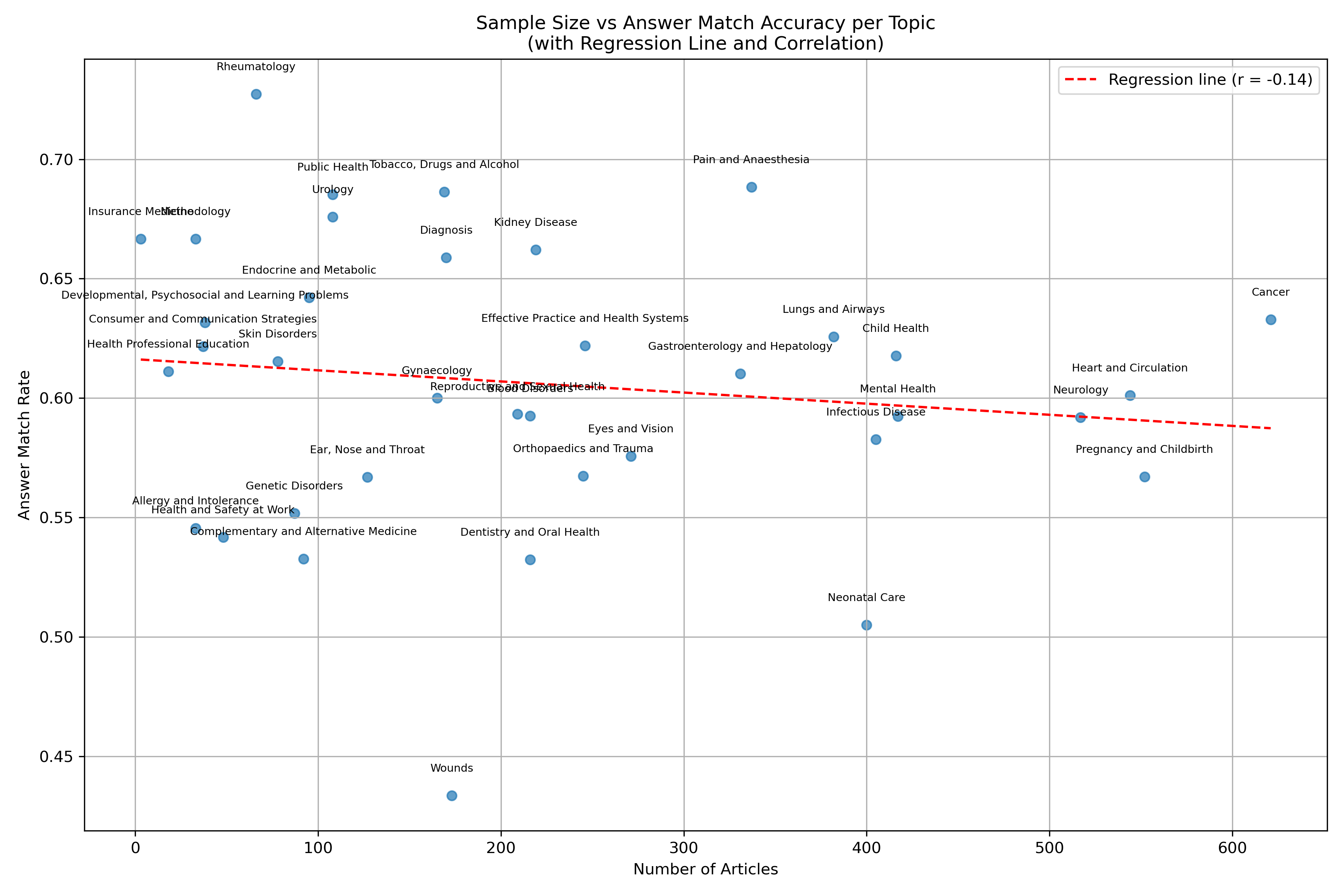}
    \caption{Accuracy for GPT-4o-mini question and answering stratified by medical domains.}
    \label{fig:topic_match}
\end{figure}

To complement the main results, we provide additional diagnostics for GPT-4o-mini. 
Figure~\ref{fig:confusion-matrix} shows confusion matrices for the answer prediction and 
discrepancy detection tasks, highlighting systematic misclassifications (e.g., frequent errors on 
``No evidence'' and ``Yes'' discrepancy cases). Table~\ref{tab:classification-report-mini} reports the 
corresponding precision, recall, and F1-scores by class, offering a more granular view of the model 
performance beyond overall accuracy. There is no evidence suggesting that accuracy changes by the number of papers going into the systematic review datasets (see Figure~\ref{fig:topic_match}).

\begin{table}[htbp]
\centering
\caption{Classification report for GPT-4o-mini ($N=8529$).}
\label{tab:classification-report-mini}
\resizebox{0.65\columnwidth}{!}{%
\begin{tabular}{lcccc}
\toprule
\textbf{Answer (Class)} & \textbf{Precision} & \textbf{Recall} & \textbf{F1-score} & \textbf{Support} \\
\hline
No          & 0.49 & 0.40 & 0.44 & 2309 \\
No Evidence & 0.19 & 0.17 & 0.18 & 1054 \\
Yes         & 0.71 & 0.78 & 0.74 & 5168 \\
\bottomrule
\end{tabular}
}
\vspace{0.5em}
\resizebox{0.65\columnwidth}{!}{%
\begin{tabular}{lcccc}
\toprule
\textbf{Discrepancy (Class)} & \textbf{Precision} & \textbf{Recall} & \textbf{F1-score} & \textbf{Support} \\
\hline
Missing & 0.52 & 0.17 & 0.26 & 2756 \\
No      & 0.69 & 0.78 & 0.73 & 5545 \\
Yes     & 0.05 & 0.30 & 0.09 & 230 \\
\bottomrule
\end{tabular}
}
\end{table}

\end{document}